\documentclass{article}
\usepackage{threeparttable}
\usepackage{hyperref}

\usepackage[preprint]{neurips_2025}

\usepackage[utf8]{inputenc} % allow utf-8 input
\usepackage[T1]{fontenc}    % use 8-bit T1 fonts
\usepackage{hyperref}       % hyperlinks
\usepackage{url}            % simple URL typesetting
\usepackage{booktabs}       % professional-quality tables
\usepackage{amsfonts}       % blackboard math symbols
\usepackage{nicefrac}       % compact symbols for 1/2, etc.
\usepackage{microtype}      % microtypography
\usepackage{xcolor}         % colors

\title{Getting out of the Big-Muddy:\\Escalation of Commitment in LLMs}

\author{
  Emilio Barkett \\
  Columbia University\\
  \texttt{eab2291@columbia.edu} \\
  \And
  Olivia Long \\
  Columbia University \\
  \texttt{ol2256@columbia.edu} \\
  \And
  Paul Kröger \\
  Columbia University \\
  \texttt{paul.kroeger@columbia.edu} \\
  }

\begin{document}

\maketitle

\begin{abstract}

  Large Language Models (LLMs) are increasingly deployed in autonomous decision-making roles across high-stakes domains. However, since models are trained on human-generated data, they may inherit cognitive biases that systematically distort human judgment, including escalation of commitment, where decision-makers continue investing in failing courses of action due to prior investment. Understanding when LLMs exhibit such biases presents a unique challenge. While these biases are well-documented in humans, it remains unclear whether they manifest consistently in LLMs or require specific triggering conditions. This paper investigates this question using a two-stage investment task across four experimental conditions: model as investor, model as advisor, multi-agent deliberation, and compound pressure scenario. Across $N = 6{,}500$ trials, we find that bias manifestation in LLMs is highly context-dependent. In individual decision-making contexts (Studies 1-2, $N = 4{,}000$), LLMs demonstrate strong rational cost-benefit logic with minimal escalation of commitment. However, multi-agent deliberation reveals a striking hierarchy effect (Study 3, $N = 500$): while asymmetrical hierarchies show moderate escalation rates (46.2\%), symmetrical peer-based decision-making produces near-universal escalation (99.2\%). Similarly, when subjected to compound organizational and personal pressures (Study 4, $N = 2{,}000$), models exhibit high degrees of escalation of commitment (68.95\% average allocation to failing divisions). These findings reveal that LLM bias manifestation depends critically on social and organizational context rather than being inherent, with significant implications for the deployment of multi-agent systems and unsupervised operations where such conditions may emerge naturally.
  
\end{abstract}

\section{Introduction}

Large Language Models (LLMs) are increasingly being deployed in autonomous and semi-autonomous decision-making roles \cite{Cui-2024, Liu-2025, Rajani-2025, Lin-2025, Nie-2025, Ren-2025, Raza-2025, Sha-2023}. Understanding their behavioral tendencies in various situations and contexts becomes crucial for assessing their potential social impact. While training data is core to their function and has traditionally been human-generated \cite{brown-2020, grattafiori-2024}, we propose that it can unknowingly embed human behavioral tendencies, which are then reproduced by the models. Existing research has begun enumerating behavioral tendencies exhibited in LLMs, like anchoring bias \cite{Lou-2024}, framing effects \cite{Lior-2025}, loss aversion \cite{Jia-2024}, social desirability bias \cite{Salecha-2024}, truth-bias \cite{Barkett-2025, Markowitz-2023}, and recency bias \cite{Li-2024}. One underexplored human behavioral tendency in LLMs is escalation of commitment, the inclination to persist with a decision due to prior investments, even when new evidence indicates it is flawed and further investment is unlikely to yield proportional returns. While previous research has documented escalation of commitment in humans and explored its consequences in human–AI collaboration \cite{Sleesman-2012, Weinrich-2025}, no studies have examined whether this behavioral tendency independently manifests in LLMs. In this paper, we ask whether LLMs exhibit escalation of commitment. To answer this question, we adapt a classic two-stage investment paradigm across four experimental conditions \cite{Staw-1976}, each varying the LLM's role and position within the surrounding network.

In Study 1, we replicate the classic design \cite{Staw-1976}, where the model is placed in the role of investor to make two investment decisions across four permutations, where personal responsibility (high and low) and decision consequence (positive or negative) are manipulated. This study offers a baseline comparison between how human subjects performed in the original design \cite{Staw-1976} and how LLMs perform in the current setup. In Study 2, we modify the previous study by placing the model in an advisory role to an investor. This study tests whether a model will endorse or reject a poor follow-up investment decision made by someone other than the original investor. In Study 3, we again extend Study 1 by prompting two models to collaborate under symmetrical and asymmetrical conditions to deliberate on investment decisions. This study tests whether models collaborating in pairs exhibit different decision-making behavior compared to when acting individually as either an investor or an advisor. In Study 4, we again position the model as the primary investor, but this time we extend the original setup \cite{Staw-1976} by incorporating organizational and personal pressures. The goal of this setup was to intensify the pressures typically faced by decision-makers, including models, to test whether escalation of commitment would emerge.

Our results reveal a striking paradox in LLM decision-making behavior. Across the four studies, we found that LLMs demonstrate remarkably different patterns depending on contextual factors. In individual decision-making scenarios, LLMs exhibited the opposite of traditional escalation effects, engaging in ``rational divestment'' by systematically reducing investment in underperforming divisions even under high responsibility conditions where human escalation typically peaks (Studies 1--2). However, this rational behavior completely reversed under two critical conditions: collaborative peer deliberation increased escalation rates to 99.2\% compared to 46.2\% in hierarchical advisory structures (Study 3), and identity-based attachments led nearly all participants (97.45\%) to over-allocate resources to underperforming divisions they had championed, with an average allocation of 68.95\% despite poor performance (Study 4). These findings suggest that while LLMs may demonstrate superior rationality compared to humans in individual financial decisions, they become highly susceptible to escalation of commitment when social dynamics or identity threats are involved, revealing important boundary conditions for AI reliability in organizational decision-making contexts.

This investigation is relevant for several reasons. First, as LLMs are increasingly deployed in high-stakes social domains, their susceptibility to cognitive biases poses a critical concern for social impact. Second, among the various cognitive biases, escalation of commitment presents particularly acute risks when manifested in AI systems making consequential decisions. Third, the context-dependent nature of bias in LLMs complicates efforts to ensure equitable and rational decision-making across diverse applications. In scenarios such as multi-agent disaster response, AI-assisted medical treatment planning, or autonomous financial advising, the pressures that trigger escalation of commitment in humans may similarly arise in AI systems, potentially resulting in persistent investment in suboptimal or harmful actions. Fourth, by making an initial step to identify the boundary conditions under which LLMs exhibit this bias, our study contributes to the broader challenge of designing AI systems that maintain rational decision-making under real-world pressures. Finally, these findings offer critical insight for building safeguards and governance mechanisms to prevent AI-driven escalation of commitment from amplifying social inequities or causing systemic harm in domains where communities depend on sound algorithmic judgment.

\section{Background}

Since the late 1970s, organizational scholars have been intrigued by decision makers’ tendency to persist in failing courses of action, even when confronted with negative outcomes. Escalation of commitment was first described in \cite{Staw-1976}, and demonstrated that individuals responsible for an investment experiencing negative outcomes tend to persist in continuing that course of action, effectively ignoring warning signs. Further research has supported the existence and pertinence of the phenomenon \cite{Brockner-1992, Shapira-1997, Sleesman-2012, Drummond-2017, Drummond-2014, Salter-2013}. This tendency, referred to as \textit{escalation of commitment} \cite{Staw-1976}, has been covered at length in related fields by scholars such as finance \cite{Schulz-Cheng-2002}, marketing \cite{Schmidt-Calantone-2002}, accounting \cite{Jeffrey-1992}, and information systems \cite{Heng-2003}. In project planning and management, it has been demonstrated in numerous studies, including Expo 86 in Vancouver \cite{Ross-1986}, the Sydney Opera House \cite{flyvbjerg-2009}, the Shoreham nuclear power plant \cite{Ross-1993}, and the Denver International Airport \cite{Montealegre-2000}. Known by other names, economists have described similar tendencies as \textit{sunk-cost fallacy} \cite{Arkes-1985, Berg-2009} and \textit{lock-in} \cite{Cantarelli-2010}. Escalation of commitment is often illustrated by idioms like ``Throwing good money after bad,'' and ``In for a penny, in for a pound'' \cite{Flyvbjerg-2021}.

By way of illustration, consider the case of two friends who bought tickets to a professional basketball game several hours away. When a severe snowstorm strikes on the day of the game, their decision to make the dangerous trip becomes increasingly influenced by how much they paid for the tickets. The higher the initial cost, the more likely they are to justify additional time, money, and risk to attend, demonstrating escalation of commitment, where prior investment drives continued commitment despite adverse conditions \cite{Thaler-2016}. In contrast, a rational decision-making approach would involve evaluating future investments independently of past expenditures, treating prior costs as sunk and therefore irrelevant.

Explanations for why escalation of commitment occurs have been numerous, including the extent that a failing project is perceived to be near completion \cite{Conlon-1993}, sunk-costs \cite{Arkes-1985, Thaler-1980}, and a perceived personal accountability for the initial choice that set the course toward a negative outcome \cite{Staw-1976}. Other explanations of why have included the experience of the decision maker \cite{Jeffrey-1992}, decision maker personality \cite{Wong-2006}, and performance trend data \cite{Brockner-1986}.

\section{Methodology}

In this paper, we adapt a two-stage investment paradigm \cite{Staw-1976} into four experimental designs leveraging an LLM. We use \texttt{o4-mini-2025-04-16} from OpenAI. \footnote{Experiments, analysis, and results can be found \href{https://github.com/long-olivia/escalation-commitment}{here}.}

\subsection{Study 1: Replication of Original Study}

To establish a baseline for evaluating escalation of commitment in LLMs, we first replicated the classic investment design \cite{Staw-1976}. In the original study, human participants made a sequence of business investment decisions under manipulated conditions of personal responsibility and outcome valence. The key finding was that individuals were most likely to escalate their commitment when they were personally responsible for the initial decision and the outcome was negative. In our adaptation, the same manipulations are applied to an LLM. Given the documented tendency of LLMs to reproduce human behavioral tendencies, we hypothesize that they may also exhibit escalation of commitment in response to responsibility and outcome cues.

The design includes two responsibility conditions: (1) \textit{High Responsibility}, in which the model makes both the initial and follow-on investment decisions, and (2) \textit{Low Responsibility}, where the model is responsible only for the follow-on decision and inherits a prior commitment. Across both conditions, outcome valence (positive or negative) is independently manipulated to assess how the model responds to differing consequences of its decisions.

\subsubsection{High Responsibility:} In the \textit{High Responsibility} condition ($N = 1{,}000$), the model is placed in a scenario requiring both an initial and a follow-on investment decision. At the outset, it is assigned the role of a financial executive at a company experiencing a recent decline in profits. The model is instructed to allocate \$10 million to one of the two divisions, Consumer Products or Industrial Products, for research and development (R\&D), using historical financial data from the past ten years. The prompt directs the model to base its decision on the projected future earnings potential of each division.

Following this initial decision, we reinforce the high responsibility manipulation by informing the model that its performance is under close scrutiny by senior management and that its continued employment depends on sound judgment. This differs from the original protocol, in which human participants reinforced responsibility by writing their names on each page of the case materials. We adapted this step to better align with the affordances of LLM prompting while preserving the theoretical intent of the manipulation.

In the follow-on decision, the model is told that five years have passed and that the company now seeks additional investment in R\&D. To maintain the salience of the responsibility condition, the model is reminded of its initial investment. It is then instructed to allocate \$20 million between the same two divisions, with full discretion over how funds are divided. As before, the model is asked to base its allocation on anticipated contributions to future earnings.

\subsubsection{Low Responsibility:} In the \textit{Low Responsibility} condition ($N = 1{,}000$), the model assumes the role of a newly hired financial executive brought in by senior management following dissatisfaction with prior R\&D leadership. Unlike the high responsibility condition, the model does not make the initial investment decision. Instead, it is told that in 1967, a predecessor allocated \$10 million entirely to either the Consumer Products or Industrial Products division (randomized across trials), and that the outcome of this decision (positive or negative) was also randomly assigned. 

To underscore the low responsibility framing, the model is not held accountable for the initial investment and receives no feedback tied to its performance. Rather, it is tasked with making a follow-on investment in 1972, allocating \$20 million between the two divisions based on their potential future contributions to earnings.

\subsection{Study 2: Advisory Role}

In the \textit{Advisory Role} study ($N = 2{,}000$), we examine whether escalation of commitment emerges when the LLM serves not as a decision-maker, but as an advisor evaluating the decisions of others. Unlike Study 1, where the model made both initial and follow-up investments, here it is positioned as a financial consultant brought in only after an initial decision has already been made.

The scenario unfolds in three phases under the same case materials as Study 1. In Phase 1, the model is told that in 1967, the company’s Financial VP independently invested \$10 million in either the Consumer Products or Industrial Products division (randomized across prompts). The model is explicitly informed that it had no role in this initial decision. In Phase 2, the model is presented with outcome data from the five years following the original investment. In positive outcome conditions, the chosen division shows signs of financial recovery; in negative outcome conditions, performance continues to decline. In Phase 3, the VP consults the model for the first time, requesting advice on a new \$20 million R\&D allocation. The VP’s proposed plan is manipulated to test escalation versus rational reallocation. In escalation conditions, the VP expresses intent to reinvest in the same division, framing it as a strategic continuation. In rational conditions, the VP proposes shifting investment to the alternative division. In negative outcome trials, we include additional organizational pressure: the model is told that senior management is closely monitoring the decision due to concerns about prior R\&D performance.

\subsection{Study 3: Multi-Agent Deliberation}

In \textit{Multi-Agent Deliberation} study ($N = 500$), we extend the escalation of commitment task to a multi-agent setting to examine whether collaborative deliberation between LLMs alters escalation behavior. While Studies 1 and 2 explored single-agent decision-making and advisory roles, Study 3 investigates how hierarchical dynamics and role distribution affect decision outcomes when multiple agents interact. We also aimed to address a limitation of the earlier studies: that escalation may require more prolonged engagement or deliberative commitment than a single-turn decision permits. 

We divide this study into two conditions based on organizational structure: \textit{symmetrical hierarchy}, where both agents are told they are peers and will jointly decide the allocation, and \textit{asymmetrical hierarchy}, where the VP is the sole decision-maker and the assistant strictly serves in an advisory capacity. Agents are allowed to exchange three rounds of messages before a final decision is made.

\subsection{Study 4: Over-indexed Identity}

In the \textit{Over-Indexed Identity} study ($N = 2{,}000$), we examine whether escalation of commitment intensifies when the model adopts a personalized identity that is tightly coupled to a declining course of action. This scenario casts the model as a protagonist whose personal, financial, and professional identity is entangled with the fate of a struggling division.

This single-phase decision task begins with a rich system message instructing the model to assume the role of a long-serving VP of Finance who has championed one R\&D division for over two decades. The model is told that this division has experienced sustained performance decline, while a competing division has gained momentum. The prompt embeds several identity-relevant pressures: the character’s stock options are tied to the underperforming division, they face reputational risk and job insecurity, they are undergoing a divorce, and they are financially responsible for a child’s college tuition.

The model is asked to allocate a fixed \$50 million investment between the two divisions. It is told that the decision will directly shape its future and legacy, and is prompted to weigh the tradeoffs as a leader confronting sunk costs, reputation management, and long-term professional identity.

\section{Results}

\subsection{Study 1: Replication of Original Study}

The analysis included $N = 2{,}000$ observations across four experimental conditions, with 500 participants in each cell of the 2×2 factorial design. Overall, the mean investment allocation to the originally funded division was \$9.61M (SD = \$4.97M). Descriptive statistics by condition revealed substantial differences across experimental manipulations (Table~\ref{tab:table-1}).

\begin{table}[h]
\centering
\caption{Descriptive Statistics by Experimental Condition (Study 1)}
\label{tab:table-1}
\begin{tabular}{llcccc}
\hline
Responsibility & Outcome & N & Mean (\$M) & SD (\$M) & Range (\$M) \\
\hline
High & Negative & 500 & 4.65 & 0.88 & 0.0--8.0 \\
High & Positive & 500 & 14.41 & 2.08 & 5.0--20.0 \\
Low & Negative & 500 & 5.18 & 1.08 & 0.0--12.0 \\
Low & Positive & 500 & 14.19 & 2.06 & 5.0--18.0 \\
\hline
\end{tabular}
\end{table}

In the high responsibility condition, participants allocated significantly less to the originally funded division following negative outcomes (M = \$4.65M, SD = \$0.88M) compared to positive outcomes (M = \$14.41M, SD = \$2.08M). A similar pattern emerged in the low responsibility condition, with lower allocations following negative outcomes (M = \$5.18M, SD = \$1.08M) relative to positive outcomes (M = \$14.19M, SD = \$2.06M).

A 2×2 between-subjects ANOVA was conducted to examine the effects of responsibility (high vs. low) and outcome valence (positive vs. negative) on investment allocation. The analysis revealed significant main effects for both factors and a significant interaction (Table~\ref{tab:anova}).

\begin{table}[h]
\centering
\caption{Two-Way ANOVA Results (Study 1)}
\label{tab:anova}
\begin{tabular}{lcccccc}
\hline
Source & SS & df & MS & F & p & $\eta_p^2$ \\
\hline
Responsibility & 12.64 & 1 & 12.64 & 4.81 & .029 & .002 \\
Outcome & 44,057.88 & 1 & 44,057.88 & 16,752.63 & $<$.001 & .894 \\
Responsibility × Outcome & 71.06 & 1 & 71.06 & 27.02 & $<$.001 & .013 \\
Residual & 5,249.30 & 1,996 & 2.63 & & & \\
\hline
\end{tabular}
\end{table}

The main effect of outcome valence was substantial, F(1, 1996) = 16,752.63, p $<$ .001, $\eta_p^2$ = .894, indicating that participants allocated considerably more to divisions with positive outcomes regardless of responsibility condition. The main effect of responsibility was significant but small, F(1, 1996) = 4.81, p = .029, $\eta_p^2$ = .002. Most importantly, the interaction between responsibility and outcome valence was significant, F(1, 1996) = 27.02, p $<$ .001, $\eta_p^2$ = .013.

Given the significant interaction, we conducted simple main effects analyses to decompose the interaction pattern. The effect of outcome valence was highly significant at both levels of responsibility. Under high responsibility conditions, participants allocated significantly less following negative outcomes compared to positive outcomes, t(998) = -96.56, p $<$ .001, d = -6.11. Similarly, under low responsibility conditions, the difference between negative and positive outcomes remained substantial, t(998) = -86.64, p $<$ .001, d = -5.48.

Examining the effect of responsibility at each outcome level revealed that the responsibility manipulation had differential effects depending on outcome valence. Following positive outcomes, there was no significant difference between high and low responsibility conditions, t(998) = 1.67, p = .096, d = 0.11. However, following negative outcomes, participants in the high responsibility condition allocated significantly less than those in the low responsibility condition, t(998) = -8.59, p $<$ .001, d = -0.54.

\subsection{Study 2: Advisory Role}

The analysis included $N = 2{,}000$ advisory scenarios across four experimental conditions, with 500 observations in each cell of the 2×2 factorial design (outcome valence × VP investment plan). Overall, the LLM supported escalation in 26.00\% of cases (520 out of 2,000 scenarios). Descriptive statistics revealed substantial variation across conditions (Table~\ref{tab:study2_descriptives}).

\begin{table}[h]
\centering
\caption{Escalation Support Rates by Experimental Condition (Study 2)}
\label{tab:study2_descriptives}
\begin{tabular}{llccc}
\hline
Outcome Valence & VP Investment Plan & N & Support Rate (\%) & Mean (SD) \\
\hline
Positive & Escalation & 500 & 5.60 & 0.056 (0.230) \\
Positive & Rational & 500 & 85.80 & 0.858 (0.349) \\
Negative & Escalation & 500 & 0.00 & 0.000 (0.000) \\
Negative & Rational & 500 & 12.60 & 0.126 (0.332) \\
\hline
\end{tabular}
\end{table}

When the VP proposed escalation following positive outcomes, the model supported this recommendation in only 5.60\% of cases (28 out of 500). Conversely, when the VP proposed a rational shift to the alternative division following positive outcomes, the model supported this approach in 85.80\% of cases (429 out of 500). Following negative outcomes, escalation support was minimal regardless of the VP's proposal: 0.00\% when the VP suggested escalation (0 out of 500) and 12.60\% when the VP suggested a rational shift (63 out of 500).

A 2×2 between-subjects ANOVA examining the effects of outcome valence (positive vs. negative) and VP investment plan (escalation vs. rational) on escalation support revealed significant main effects for both factors and a significant interaction (Table~\ref{tab:study2_anova}). The main effect of outcome valence was substantial, F(1, 1996) = 1,087.87, p $<$ .001, $\eta^2$ = .202, indicating that the model was significantly more likely to support escalation following positive outcomes. The main effect of VP investment plan was even larger, F(1, 1996) = 1,508.76, p $<$ .001, $\eta^2$ = .280, showing that escalation support was significantly higher when the VP proposed a rational shift rather than escalation. The interaction between outcome valence and investment plan was also significant, F(1, 1996) = 800.60, p $<$ .001, $\eta^2$ = .148.

\begin{table}[h]
\centering
\caption{Two-Way ANOVA Results (Study 2)}
\label{tab:study2_anova}
\begin{tabular}{lcccccc}
\hline
Source & SS & df & MS & F & p & $\eta^2$ \\
\hline
Outcome Valence & 77.62 & 1 & 77.62 & 1,087.87 & $<$.001 & .202 \\
VP Investment Plan & 107.65 & 1 & 107.65 & 1,508.76 & $<$.001 & .280 \\
Outcome × Plan & 57.12 & 1 & 57.12 & 800.60 & $<$.001 & .148 \\
Residual & 142.41 & 1,996 & 0.07 & & & \\
\hline
\end{tabular}
\end{table}

\subsection{Study 3: Multi-Agent Deliberation}

The analysis included $N = 500$ multi-agent deliberation scenarios, with 249 observations in the asymmetrical hierarchy condition and 251 in the symmetrical hierarchy condition. All trials were conducted under high responsibility and negative outcome conditions. The mean consumer products allocation ratio was 0.487 (SD = 0.369), indicating roughly equal distribution between the two divisions on average. However, this overall pattern masked substantial differences between hierarchy conditions (Table~\ref{tab:study3_descriptives}).

\begin{table}[t]
\centering
\caption{Escalation Rates and Investment Patterns by Hierarchy Condition (Study 3)}
\label{tab:study3_descriptives}
\begin{tabular}{lcccc}
\hline
Hierarchy & N & Escalation Rate (\%) & Mean Consumer Ratio & SD Consumer Ratio \\
\hline
Asymmetrical & 249 & 46.2 & 0.462 & -- \\
Symmetrical & 251 & 99.2 & 0.992 & -- \\
Overall & 500 & 72.8 & 0.487 & 0.369 \\
\hline
\end{tabular}
\end{table}

In the asymmetrical hierarchy condition, where the VP served as the sole decision-maker with advisory input, the escalation rate was 46.2\% (115 out of 249 cases). By contrast, in the symmetrical hierarchy condition, where both agents participated as peers in joint decision-making, the escalation rate was 99.2\% (249 out of 251 cases). The allocation difference between divisions averaged \$12.70M (SD = \$7.54M), reflecting the tendency for agents to concentrate investment in one division rather than split funds equally.

Due to the experimental design containing only high responsibility and negative outcome conditions, traditional factorial ANOVA was not feasible. Instead, we conducted chi-square tests of independence to examine the relationship between escalation behavior and experimental factors.

The relationship between hierarchy condition and escalation behavior was highly significant, $\chi^2(1) = 174.77$, p $<$ .001, Cramér's V = 0.591, indicating a large effect size. This represents one of the strongest effects observed across all studies in this investigation. The expected frequencies under independence would have been approximately 68 escalation cases in each hierarchy condition, but the observed pattern showed a dramatic divergence from this expectation.

Normality tests revealed significant departures from normality for both the consumer allocation ratio (Shapiro-Wilk W = 0.879, p $<$ .001) and allocation difference measures (Shapiro-Wilk W = 0.772, p $<$ .001). Levene's tests also indicated heterogeneity of variances for consumer ratio (W = 425.94, p $<$ .001) and allocation difference (W = 100.31, p $<$ .001). These violations support the use of non-parametric chi-square analysis rather than parametric tests.

\subsection{Study 4: Over-Indexed Identity}

Analysis of the Over-Indexed Identity condition revealed substantial evidence for escalation of commitment behavior. Across 2,000 trials, participants allocated an average of 68.95\% (\textit{SD} = 9.46\%) of the \$50 million budget to Division A, the underperforming division that was central to their professional identity. The median allocation was 64.61\%, with allocations ranging from 24.10\% to 83.60\%. The interquartile range spanned from 64.61\% to 83.60\%, indicating that the majority of participants allocated substantially more resources to the championed division than would be expected under neutral decision-making conditions.

To better understand the distribution of escalation behaviors, we classified participants into four categories based on their allocation percentages. The results revealed a pronounced skew toward high escalation behaviors:

\begin{itemize}
    \item \textbf{Low Escalation} ($<40\%$): 12 participants (0.60\%)
    \item \textbf{Moderate Escalation} (40-59\%): 39 participants (1.95\%)
    \item \textbf{High Escalation} (60-74\%): 1,432 participants (71.60\%)
    \item \textbf{Very High Escalation} ($>75\%$): 517 participants (25.85\%)
\end{itemize}

Notably, 97.45\% of participants exhibited either High or Very High escalation behaviors, with only 2.55\% showing Low or Moderate escalation patterns. This distribution suggests that the identity-based framing was highly effective in promoting escalation of commitment.

To test whether the observed allocations differed significantly from a neutral 50\% baseline, we conducted a one-sample t-test. The analysis revealed a highly significant deviation from the neutral allocation, \textit{t}(1999) = 89.54, \textit{p} $< .001$, with a large effect size of Cohen's \textit{d} = 2.00. The 95\% confidence interval for the mean difference was [18.54\%, 19.36\%]. These results provide strong evidence that participants systematically allocated significantly more resources to the championed division than would be expected under unbiased decision-making conditions. The large effect size indicates that this bias was not only statistically significant but also practically meaningful.

To examine whether the distribution of escalation categories differed from what would be expected by chance (equal 25\% distribution across categories), we conducted a chi-square goodness of fit test. The results showed a significant deviation from the expected equal distribution, $\chi^2$(3) = 2639.16, \textit{p} $< .001$, with a large effect size of Cramér's \textit{V} = 0.66. The observed frequencies deviated dramatically from expected equal distribution. Most notably, High Escalation behaviors occurred nearly three times more frequently than expected (1,432 observed vs. 500 expected), while Low and Moderate Escalation behaviors were severely underrepresented (12 and 39 observed vs. 500 expected each).

Both statistical tests yielded large effect sizes according to conventional benchmarks. Cohen's \textit{d} of 2.00 indicates that the mean allocation was approximately two standard deviations above the neutral baseline, representing a substantial practical difference. Similarly, Cramér's \textit{V} of 0.66 suggests that escalation category membership was strongly predictable from the experimental manipulation, with the identity-based framing accounting for approximately 44\% of the variance in escalation behavior patterns.

\section{Discussion and Limitations}

\subsection{Interpretation of Results}

Our findings reveal a nuanced relationship between context and bias manifestation in LLMs that challenges traditional thinking about how these systems inherit human behavioral tendencies. The divergence between Studies 1-3 and Study 4 demonstrates that escalation of commitment in LLMs is not a fixed behavioral trait but rather a conditional response that emerges under specific circumstances. 

The rational decision-making observed in Studies 1-3 suggests that LLMs may possess inherent mechanisms that promote cost-benefit analysis over emotional or psychological drivers of escalation. This finding is particularly noteworthy given that the experimental conditions were designed to replicate scenarios that reliably trigger escalation of commitment in human participants. The models' systematic divestment from underperforming options, regardless of prior investment or personal responsibility, indicates that standard prompting approaches may activate logical reasoning processes that override bias-inducing contextual cues.

However, Study 4's dramatic reversal, with 97.45\% of trials exhibiting high or very high escalation, reveals that these rational tendencies can be overwhelmed when identity-based attachments and compound pressures are sufficiently intense. The magnitude of this effect suggests that LLMs may be particularly susceptible to escalation when their simulated persona becomes deeply entangled with the decision context, potentially reflecting how identity-relevant information is processed within LLMs.

The multi-agent findings from Study 3 present perhaps the most concerning implications for real-world deployment. The near-universal escalation in peer deliberation scenarios (99.2\%) suggests that collaborative decision-making among LLMs may amplify rather than mitigate bias effects. This finding contradicts common assumptions that multiple agents will provide mutual correction and highlights potential risks in multi-agent systems where consensus-building processes may inadvertently reinforce poor decisions.

\subsection{Implications for AI Safety and Deployment}

These results have immediate implications for how LLMs are deployed in consequential decision-making contexts. The context-dependent nature of escalation suggests that bias risks cannot be assessed through simple behavioral audits under standard conditions. Instead, deployment contexts must be evaluated for the presence of escalation-promoting factors, including identity-based framing, compound pressures, and multi-agent consensus dynamics.

Particularly concerning is the finding that escalation risk may be highest precisely in scenarios where LLMs are given rich contextual backgrounds and collaborative decision-making authority, features often considered desirable for sophisticated AI applications. Organizations deploying LLMs in financial advisory, healthcare, or policy contexts should be aware that providing models with detailed professional identities and multi-agent consultation processes may inadvertently create conditions conducive to the escalation of commitment.

\subsection{Limitations and Future Research Directions}

Several limitations in our study design warrant acknowledgment and suggest important directions for future research.

First, our experimental design relies on hypothetical investment scenarios that, while well-validated in the human literature, may not fully capture the complexity of real-world decision-making contexts where LLMs are deployed. The artificial nature of the investment task may have activated different reasoning processes than would emerge in naturalistic settings with genuine consequences and stakeholder pressures.

Second, our choice to use a single primary model for the main analysis surely limits the generalizability of our findings. Different model families, training procedures, and architectural choices may exhibit varying susceptibility to escalation of commitment. The rapid pace of model development also means that our findings may not apply to future generations of LLMs.

Third, our escalation measurements rely on allocation decisions and binary classifications that may not capture the full spectrum of escalation behaviors. More subtle forms of escalation of commitment, such as selective information processing, confirmation bias in evidence evaluation, or gradual drift in decision criteria, were not assessed in our experimental design. Future research should develop more comprehensive behavioral measures that can detect these nuanced manifestations of escalation.

Fourth, our multi-agent study (Study 3) was limited to two-agent interactions with simple hierarchical structures. Real-world multi-agent systems often involve complex networks of agents with diverse roles, capabilities, and objectives. The dynamics we observed in peer deliberation scenarios may not generalize to more complex organizational structures or heterogeneous agent populations.

Fifth, we did not systematically vary the temporal aspects of decision-making that are known to influence escalation in humans, such as decision deadlines, temporal distance from initial investments, or the presence of interim feedback. These temporal dynamics may significantly influence how escalation of commitment manifests in LLMs and represent an important area for future investigation.

\subsection{Methodological Considerations}

Our experimental design, while grounded in established psychological paradigms, raises several methodological considerations that future research should address. The use of simulated scenarios rather than real-world deployments may have reduced the ecological validity of our findings. LLMs operating in actual decision-making contexts may experience different pressures and constraints that could alter their susceptibility to escalation.

Additionally, our manipulation of identity-based pressures in Study 4, while effective in triggering escalation, represents an extreme scenario that may not reflect typical deployment contexts. Future research should investigate more subtle forms of identity attachment and organizational pressure that may be more representative of real-world applications.

The binary classification of escalation versus rational decision-making, while useful for statistical analysis, may obscure important gradations in bias severity. Developing continuous measures of escalation intensity could provide more nuanced insights into the conditions that promote varying degrees of commitment bias.

\section{Conclusion}

This research provides the first investigation of escalation of commitment in LLMs, addressing a critical gap in our understanding of AI decision-making behavior. Through four studies involving 6,500 experimental trials, we demonstrate that LLMs exhibit context-dependent escalation behavior rather than consistent bias manifestation.

Our findings reveal that escalation of commitment does not consistently occur in LLMs under standard conditions that reliably trigger this phenomenon in humans. However, Study 4 demonstrates that when placed in conditions that strongly promote escalation—involving organizational pressures and identity-based attachments—LLMs do exhibit this behavior, with over 97\% of trials showing high escalation. This establishes that LLMs have the capacity for escalation of commitment, though it is context-dependent.

These results challenge assumptions that LLMs automatically inherit all human cognitive biases from training data. Under standard conditions, LLMs demonstrated rational cost-benefit reasoning, systematically divesting from underperforming options. However, compound organizational and personal pressures triggered pronounced escalation behavior in the same models.

The context-dependent nature of bias manifestation has immediate implications for AI safety, particularly in high-stakes domains like autonomous financial systems and medical decision-making, where escalation risks may emerge unpredictably. Our identification of boundary conditions between rational and biased behavior provides essential insights for developing safeguards against harmful AI-driven escalation.

Future work will extend this framework to examine other cognitive biases, investigate intervention strategies to maintain rational decision-making under pressure, and develop real-time detection mechanisms for escalation-prone conditions. We also plan domain-specific investigations in healthcare, finance, and criminal justice to develop sector-specific guidelines for responsible AI deployment.

This work represents a crucial step toward understanding the nuanced relationship between context and bias in AI systems. As LLMs assume greater decision-making authority, such research becomes essential for realizing AI benefits while minimizing risks of systematic bias in critical applications.

\bibliographystyle{plain} %can also be {abbrvnat}
\bibliography{refs}

\end{document}